\documentclass[10pt,twocolumn,letterpaper]{article}

\usepackage[pagenumbers]{iccv} 

\definecolor{iccvblue}{rgb}{0.21,0.49,0.74}
\usepackage[pagebackref,breaklinks,colorlinks,allcolors=iccvblue]{hyperref}
\usepackage{booktabs}
\usepackage{float}

\usepackage{siunitx}

\def\paperID{2} 
\def\confName{ICCV}
\def\confYear{2025}

\title{From Static to Dynamic: a Survey of Topology-Aware Perception \\ in Autonomous Driving}

\author{
Yixiao Chen$^{1}$\thanks{Equal Contribution}  \hspace{0.8cm}  
Ruining Yang$^{1}$\footnotemark[1]   \hspace{0.8cm}  
Xin Chen$^{2}$  \hspace{0.8cm}  Jia He${^2}$ \hspace{0.8cm} \\ 
Dongliang Xu$^{2\dag}$ \hspace{0.8cm}  Yue Yao$^2$\thanks{Corresponding authors}  \\
$^1$Sems  \hspace{0.4cm}   $^2$Shandong University 
}

\begin{document}
\maketitle
\begin{abstract}

The key to achieving autonomous driving lies in topology-aware perception, the structured understanding of the driving environment with an emphasis on lane topology and road semantics. This survey systematically reviews four core research directions under this theme: vectorized map construction, topological structure modeling, prior knowledge fusion, and language model-based perception. Across these directions, we observe a unifying trend: a paradigm shift from static, pre-built maps to dynamic, sensor-driven perception. Specifically, traditional static maps have provided semantic context for autonomous systems. However, they are costly to construct, difficult to update in real time, and lack generalization across regions, limiting their scalability. In contrast, dynamic representations leverage on-board sensor data for real-time map construction and topology reasoning. Each of the four research directions contributes to this shift through compact spatial modeling, semantic relational reasoning, robust domain knowledge integration, and multimodal scene understanding powered by pre-trained language models. Together, they pave the way for more adaptive, scalable, and explainable autonomous driving systems.

\end{abstract}

\section{Introduction}
\label{sec:intro}

The ultimate goal of autonomous driving is to achieve safe, efficient and robust Autonomous Driving (AD) in an open, dynamic real world. Achieving this goal depends on a structured understanding of the road environment, including perception and detection of surrounding obstacles. In most structural perception tasks, lane topology reasoning plays a vital role~\cite{wang2023openlane,wu2023topomlp,peng2024lanegraph2seq}. It requires the system to not only recognize the geometry of lanes, but also understand the connection relationship between lanes, semantic attributes, and interaction with traffic control elements. This ability to model road topology not only directly supports the operation of path planning and behavior prediction modules, but also is the basic guarantee for the system's generalization ability and safety~\cite{liu2023vectormapnet,ben2022toponet}.

\begin{figure*}[t]
\begin{center}
	\includegraphics[width=1\linewidth]{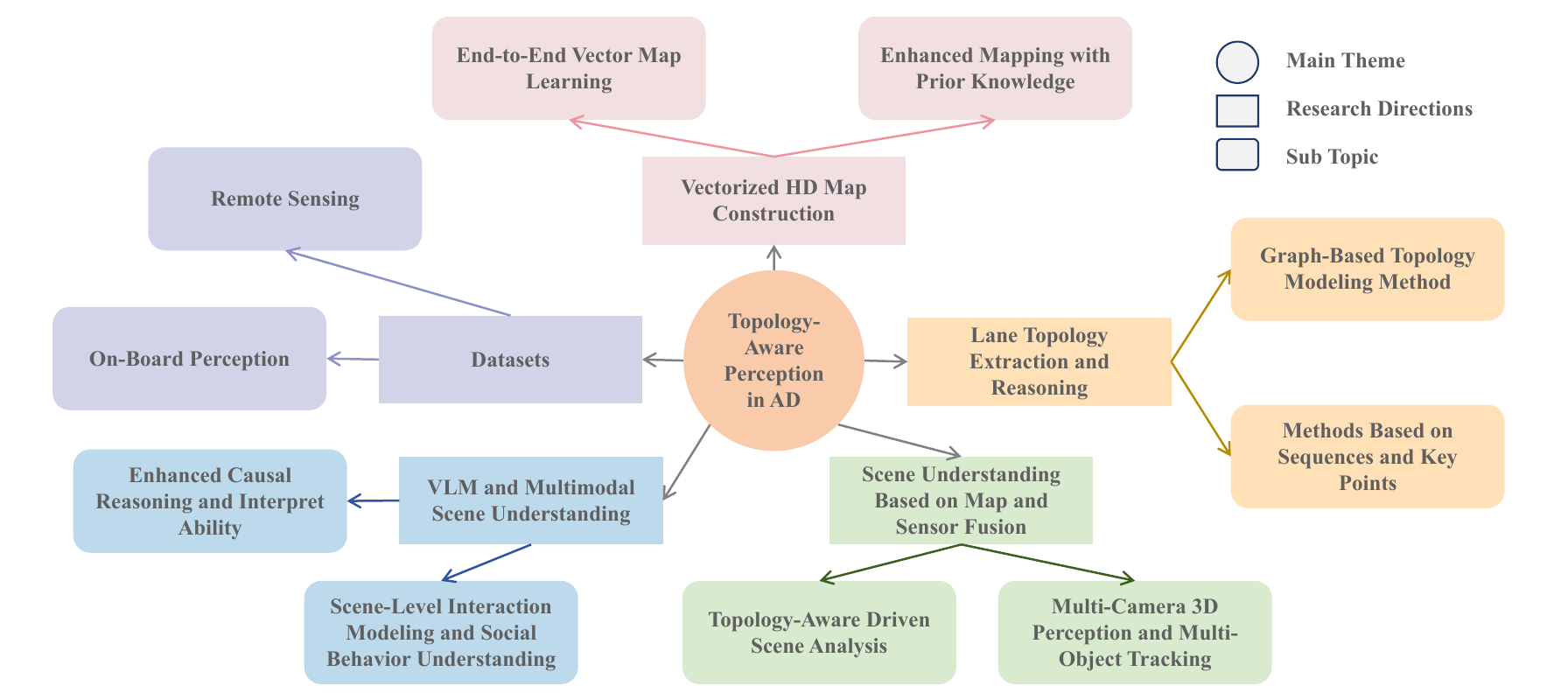}
\end{center}
\vspace{-2em}
\caption{
Survey framework. This paper systematically covers the core topics in the field of topology perception in autonomous driving. Specifically, this paper focuses on four research directions: vectorized HD map construction, lane topology extraction and reasoning, scene understanding based on map and sensor fusion, and vlm and multimodal scene understanding. Each direction is further analyzed with at least two sub topics.
 }
\label{fig:structure}
\end{figure*}

High-Definition (HD) maps have long been used to provide structured environmental information for autonomous driving~\cite{li2022hdmapnet,chang2019argoverse,huang2018apolloscape}. Their high accuracy and structured characteristics make them perform well in known environments. However, their limitations are gradually becoming apparent in the process of deployment at the city level or even across regions. The construction of high-definition maps usually relies on professional surveying and mapping equipment and a large amount of manual annotation, which is costly~\cite{zhang2014loam,zhou2022cross,hu2021fiery}. Meanwhile, its static characteristics also make it difficult to adapt to dynamic scenes such as road construction and traffic changes~\cite{yang2025histrackmap,shan2018lego,shan2020lio}. In addition, the system's high reliance on prior maps also limits its generalization ability in new environments or rare scenarios~\cite{li2025generative,prakash2021multi,xiao2020multimodal}.

To overcome these limitations, an increasing amount of research has begun to turn to online mapping, exploring real-time mapping methods that do not rely on prior maps~\cite{hu2021fiery,li2022hdmapnet,luo2024augmenting,liao2022maptr}. This shift has spawned a series of algorithmic systems designed to use on-board sensors such as cameras, lidar, and millimeter-wave radar to build semantic maps and topological structures in real time, promoting a fundamental change in the map construction paradigm.

\begin{figure*}[t]
\begin{center}
	\includegraphics[width=1\linewidth]{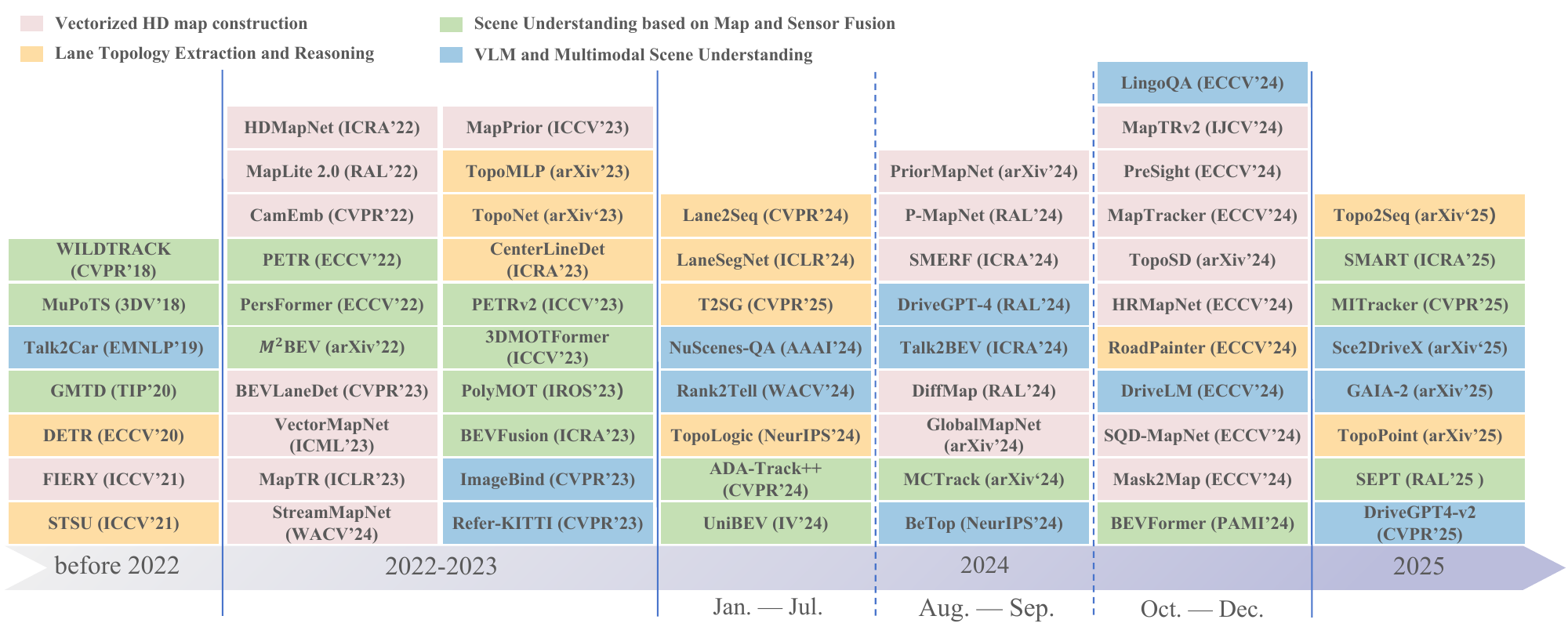}
\end{center}
\vspace{-2em}
\caption{
Following Fig.~\ref{fig:intro}, we show representive works on topology-aware perception in autonomous driving. 
These papers are sorted by publication date, and different color boxes represent different research topics, such as vectorized high-definition map construction, lane topology extraction and inference, scene understanding based on map and sensor fusion, and visual language models and multimodal scene understanding.
 }
\label{fig:intro}
\end{figure*}

In recent years, significant progress has been made in the areas of vectorized map construction, topological structure modeling, prior knowledge fusion, and language-based perception enhancement. In this survey, shown in Fig.~\ref{fig:structure} and Fig.~\ref{fig:intro}, we provide a systematic review of the latest research in map construction and topological reasoning for autonomous driving, with a particular emphasis on the evolutionary trajectory from traditional perception methods to vectorized mapping, integration of multi-source priors, and language-driven scene understanding. Our main contributions are as follows:
\begin{itemize}
  \item To the best of our knowledge, this is the first systematic review of the four major directions of topology-aware perception, \ie, vectorized map construction, topological reasoning, prior knowledge fusion, and language model perception, filling the gap in the current field that lacks a unified perspective and cross-modal thinking.
  
  \item We divide the existing research methods into four paradigms and conduct an in-depth analysis from the perspectives of mapping process, model architecture, semantic understanding ability, and multimodal fusion mechanism. We also summarize the representative work and technical evolution trends of various methods.
  
  \item We put forward the key challenges currently faced, including the difficulty of topological consistency modeling, the real-time updating of maps, and the lack of a unified expression framework for cross-modal perception. We further explore potential paths for the future development of autonomous driving systems towards generalization, explainability, and interactivity.
\end{itemize}

\section{Vectorized HD Map Construction}
\label{sec:vectorized_hd_map}
The key to moving towards online map construction is to fundamentally redefine the way road features are represented. Most early methods borrowed semantic segmentation technology to model maps as rasterized semantic labels in a bird's-eye view (BEV). However, as task requirements increase, research has gradually moved towards more efficient and structured vector representations. This section reviews the evolution of raster to vector mapping, focusing on the development of end-to-end polyline prediction methods, and explores how to combine prior knowledge to improve the accuracy and generalization of the system.

\subsection{End-to-End Vector Map Learning}

Early online map construction methods mainly adopted rasterized strategies. This type of method extracts BEV representations from multi-view cameras and performs semantic segmentation on BEV space to construct the spatial distribution of road structures and elements~\cite{philion2020lift,li2023bevdepth}. For example, HDMapNet~\cite{li2022hdmapnet} uses a multi-branch network structure to perform pixel-level predictions on map elements such as lane lines, crosswalks, and parking lines, creating a precedent for end-to-end online high-precision map perception. Similar rasterization methods include BEVLaneDet~\cite{wang2023bev} and Persformer~\cite{chen2022persformer}, which model the three-dimensional lane detection problem as a segmentation task on BEV images, further improving the accuracy of structural perception. FIERY~\cite{hu2021fiery} uses simplified BEV representation for instant segmentation, which has strong real-time performance; while CamEmb~\cite{zhou2022cross} abandons the traditional method's reliance on camera calibration parameters, and directly predicts BEV semantic masks from image features through learning-based mapping, improving cross-platform generalization capabilities.

Although such methods have significant advantages in training stability and multimodal fusion, and have been widely used in the early days, their output is usually dense pixel distribution, which is difficult to be directly used by downstream modules such as path planning and control decision-making. These modules usually rely on structured vector map inputs, so complex post-processing steps, such as skeleton extraction and curve fitting are required to convert semantic masks into vector polylines, which are often non-differentiable and prone to topological errors and geometric noise.

In order to solve this limitation, research has gradually turned to end-to-end vector mapping frameworks. Representative work VectorMapNet~\cite{liu2023vectormapnet} first proposed to directly generate sparse multi-segment polylines from images. This method adopts a coarse-to-fine two-stage strategy: first, the area where the map element is located is detected, and then the polyline vertices are generated point by point through autoregression, realizing the complete vector mapping process. However, its sequential prediction mechanism leads to low inference efficiency, and there is a problem of error accumulation when dealing with situations such as path ambiguity or endpoint uncertainty.

To overcome the above problems, MapTR~\cite{liao2022maptr} proposed a parallel mapping method based on Transformer. This method models map elements as unordered point sets, avoids dependence on generation order, and regresses all key points simultaneously in a single-stage framework, significantly improving efficiency and robustness. Its upgraded version MapTRv2~\cite{liao2024maptrv2} further introduces a decoupled self-attention mechanism and advanced training strategies to achieve better topology preservation and deployment performance.

On this basis, the researchers further explored key issues such as temporal consistency, long-distance perception and prior fusion. For example, StreamMapNet~\cite{yuan2024streammapnet} proposed a streaming mapping framework, introduced a cross-frame query propagation mechanism and a BEV fusion module, and effectively improved the consistency of long-term mapping. DiffMap~\cite{jia2024diffmap} used a diffusion model to build high-quality maps of continuous scenes; PriorMapNet~\cite{wang2024priormapnet} enhanced the generalization of mapping by aligning language and visual priors; and P-MapNet~\cite{jiang2024p} combined a multi-scale attention mechanism with scene priors to significantly improve the accuracy of long-distance map reconstruction.

\subsection{Enhanced Mapping with Prior Knowledge}

Since purely sensor-driven online map building systems are susceptible to interference from factors such as blind spots, occlusions, dynamic target occlusions, and bad weather, in recent years, more and more research works have tried to introduce map priors to enhance the stability and perception range of the system, especially in long-distance mapping and structure completion tasks.

A mainstream strategy is to use standard definition maps (SD maps) that are widely available but have limited accuracy. For example, MapLite 2.0~\cite{ort2022maplite} and TopoSD~\cite{yang2024toposd} show that the high-level topological structure provided by SD maps can be used as an effective perception constraint. TopoSD encodes SD map elements into spatial geometric structures and instance semantic tags, and integrates them into BEV representation, significantly enhancing the lane geometry and connectivity modeling capabilities in complex scenarios such as OpenLane-V2~\cite{wang2023openlane}. Similarly, SMERF~\cite{luo2024augmenting} directly integrates SD Map information for real-time map prediction, and P-MapNet~\cite{jiang2024p} further focuses on the weakly aligned SD Map skeleton through the attention mechanism, and learns the prior distribution of typical maps with the help of masked autoencoders, thereby effectively correcting unnatural prediction shapes and significantly enhancing long-distance map perception.

In addition to SD Map, satellite images are also used as a well-structured prior input to provide a top-down semantic perspective. Works such as Mapprior~\cite{zhu2023mapprior} and PreSight~\cite{yuan2024presight} explored combining satellite images with generative models such as neural radiance fields (NeRF) to construct map distributions with consistent structure and realistic textures, providing new ideas for robust map prediction under extreme conditions.

At the same time, there are also methods to model cross-frame map evolution from the perspective of temporal consistency. StreamMapNet~\cite{yuan2024streammapnet} is the first framework to propose a streaming map construction mechanism. By introducing cross-frame query propagation and multi-point attention mechanisms, it improves the coherence and stability in complex traffic scenarios. Furthermore, SQD-MapNet~\cite{wang2024stream} proposed a streaming query denoising strategy to correct historical features for map construction; while MapTracker~\cite{chen2024maptracker} transforms map construction into a multi-instance target tracking task, using latent space memory to achieve time-consistent map updates. GlobalMapNet~\cite{shi2024globalmapnet} starts from a global perspective and introduces the Map-NMS strategy for redundancy removal to ensure that the final generated vector map has a high overall topological accuracy.

In addition, some work focuses on the raster-to-vector conversion mechanism, such as Mask2Map~\cite{choi2024mask2map}, which uses a two-stage architecture to first generate a rasterized map through a semantic segmentation network, and then uses a mask-driven network for vectorization refinement. HRMapNet~\cite{zhang2024enhancing} uses city-scale historical grid maps and vehicle positioning information to perform spatial prior enhancement on the current scene to improve the accuracy and completeness of structural prediction.
\section{Lane Topology Extraction and Reasoning}
\label{sec:3_topo_reasoning}

The perception of map elements is only the basis for understanding traffic scenes. To achieve explainable and structured scene understanding, it is necessary to further reason about the spatial connection relationship between road elements. Especially in complex areas such as intersections, forks and roundabouts, geometric perception is often not enough to support behavior planning. Therefore, lane topology reasoning, as a key step in building high-order map semantics, is receiving widespread attention. Its core task is to determine the connection relationship between lanes and restore a structurally complete topological map. 

\subsection{Graph-Based Topology Modeling Method}
Graph structure modeling is currently the most widely used method in lane topology reasoning. Its core idea is to model each lane as a node in the graph, and the connection relationship between different lanes as an edge in the graph, so as to transform the complex road topology into a graph with a clear structure. 
These methods generally follow the paradigm of detection first, then reasoning: first extract geometric information through a high-precision lane detection module, and then reason about the connectivity between nodes based on spatial structure or semantic features. However, there are significant differences among the methods in terms of reasoning module design and feature modeling strategies.

TopoMLP~\cite{wu2023topomlp} is a typical lightweight graph classification method. This method first constructs candidate lane pairs as edges in the graph, and then uses a MLP to determine whether they are connected. The advantages of TopoMLP are simple structure and stable training, but its performance is limited by the quality of front-end detection and is sensitive to input spatial errors. It is worth noting that PETR~\cite{liu2022petr} is used as the front-end detector in TopoMLP to perceive the center line of the lane.

To improve robustness, TopoLogic~\cite{fu2024topologic} introduces an explicit structure modeling mechanism based on the above framework. It proposes to combine two types of signals, the Euclidean distance of endpoints and the similarity of semantic features, to improve the adaptability to perception errors and occlusions, especially in real scenes with higher prediction stability.

TopoNet~\cite{ben2022toponet} is an early baseline in the field of topological learning. It is the first to use graph neural networks to transmit information on topological maps, supporting the modeling of structures including intersections and traffic lights. CenterLineDet~\cite{xu2023centerlinedet} and TopoNet both regard lane lines as vertices in the graph and update lane representations based on graph neural networks to build semantically rich topological structure graphs.

LaneSegNet~\cite{li2023lanesegnet} introduces a ``lane attention'' mechanism to enhance lane segment perception under BEV representation, providing more stable support for the perception basis of graph structure. SMERF~\cite{luo2024augmenting} uses the SDMap as one of the perception inputs, which effectively supplements the perception loss when the structural information is sparse or severely occluded.

Further research also attempts to improve the structure generation strategy. For example, STSU~\cite{can2021structured} borrows the idea of DETR~\cite{carion2020end} and introduces a minimum loop query mechanism to optimize lane segment sorting and connection judgment; Can et al.~\cite{can2022topology} also improves the connection accuracy between multiple lane segments through topological consistency constraints.

In addition, T2SG~\cite{lv2025t2sg} constructs a more fine-grained topological graph, uniformly modeling lane segments, control signals, lane attributes, etc. as graph nodes, explicitly expressing structural relationships such as "lane-signal light" and "main lane-control lane", which is suitable for complex traffic network modeling.

\subsection{Methods Based on Sequences and Key Points}

In order to further improve the efficiency of reasoning and reduce the dependence on the quality of graph construction, a series of problems have emerged in recent years that model topology extraction tasks as sequence generation or key point location. Such methods are usually more flexible and suitable for urban scenes with complex structures and open topologies.

Lane2Seq~\cite{zhou2024lane2seq} and Topo2Seq~\cite{yang2025topo2seq} are representative sequence modeling methods. Inspired by the Seq2Seq~\cite{sutskever2014sequence} in natural language processing, the topological structure is represented as a sequence of topological units and the Transformer architecture is used for end-to-end generation, eliminating the need for complex graph structure encoding and post-processing.

RoadPainter~\cite{ma2024roadpainter} and TopoPoint~\cite{fu2025topopoint} explore connectivity reasoning mechanisms with key points as the core. RoadPainter locates lane connection key points through BEV masks and determines whether they are connected based on the geometric relationship between point pairs. TopoPoint goes a step further by modeling endpoints as independent entities and integrating the geometric and semantic relationship information between endpoints for connection prediction, which significantly improves the robustness of the model to perceptual anomalies such as occlusion, offset, and breakage.

In addition, some methods also integrate temporal information to enhance the temporal consistency of topological mapping. For example, StreamMapNet~\cite{yuan2024streammapnet} introduces a temporal mapping mechanism to improve the consistency and accuracy of continuous frame mapping through query propagation or stream query denoising strategies.
\section{Scene Understanding Based on Map and Sensor Fusion}
\label{sec:4_scene_understanding}

High-level topological mapping and behavior reasoning tasks essentially rely on robust and sophisticated underlying perception capabilities. On-board sensor perception not only provides input for map element extraction, but also provides structural support for dynamic object modeling and traffic scene analysis. This section focuses on high-level perception tasks such as multi-camera 3D perception, multi-target tracking, map element detection, and fusion of topological information, and reviews current mainstream technologies and their role in autonomous driving.

\subsection{Multi-Camera 3D Perception and Multi-Object Tracking}

In vision-driven autonomous driving systems, how to reconstruct consistent 3D semantic scenes from multiple 2D perspectives is a key challenge for achieving BEV perception and map construction. PETR~\cite{liu2022petr} first introduced position encoding, integrating 3D information into image features, breaking the reliance on BEV explicit mapping. Subsequently, PETRv2~\cite{liu2023petrv2} added temporal modeling and task-customized query mechanisms, unified support for 3D detection, BEV segmentation, and lane line detection, taking into account both accuracy and efficiency.

To better integrate multi-view information, methods such as BEVFusion~\cite{liu2023bevfusion},  \( M^2 \)BEV~\cite{xie2022m} and BEVFormer~\cite{li2024bevformer} complete feature projection and integration in the detection stage; UniBEV~\cite{wang2024unibev} further supports cross-height modeling and improves occlusion and distortion robustness.

In terms of dynamic target modeling, 3DMOTFormer~\cite{ding20233dmotformer} uses graph transformers to achieve context modeling between detection and trajectory, and introduces an online adaptive training strategy. ADA-Track++~\cite{ding2024ada} achieves detection-tracking collaborative optimization through a learnable association module, significantly improving tracking accuracy in occluded and dense scenes. MCTrack~\cite{wang2024mctrack} and PolyMOT~\cite{li2023poly} proposed a multi-camera spatial consistency modeling mechanism to enhance the cross-view target representation capability. MITracker~\cite{xu2025mitracker} can track any object in any video frame of any view and any length.

In addition, multi-view tracking research also exposes the limitations of datasets. While existing datasets such as WILDTRACK~\cite{chavdarova2018wildtrack} and MuPoTS~\cite{mehta2018single} primarily emphasize pedestrian targets, efforts to broaden the scope—such as in GMTD~\cite{wu2020visual}, which includes a more diverse set of object categories—are constrained by the dataset's relatively small scale. Promoting larger-scale, open-class multi-view datasets is still an important direction.

In terms of map element perception, LaneSegNet~\cite{li2023lanesegnet} uses a multi-branch structure to perceive the boundaries of different lanes; CenterLineDet~\cite{xu2023centerlinedet} and SEPT~\cite{pei2025sept} integrate map priors and perception features to effectively improve the perception accuracy of centerlines and topological structures, providing key input for subsequent topological mapping.

\subsection{Topology-Aware Driven Scene Analysis}

To achieve a more comprehensive and semantically rich scene understanding, some studies have explicitly introduced topological structures into the perception process and used structural constraints to enhance the expression and understanding capabilities of BEV. The Structured BEV method~\cite{can2021structured,rong2024driving} models lanes and traffic elements as structured objects, achieving an effective transition from perception to scene graphs, and improving consistency and generalization performance in complex scenes. PersFormer~\cite{chen2022persformer} optimizes lane geometry expression through perspective transformation, alleviates image scale distortion, and provides accurate input for subsequent topological mapping.

In addition, map priors not only help map construction but can also be used for perception enhancement. SMART~\cite{ye2025smart} proposed a transferable structural prior module to introduce structural knowledge from SDMap or satellite images into mainstream perception systems, significantly improving robustness under occlusion and complex environments. Other methods~\cite{luo2024augmenting,zhang2024enhancing,xiong2023neural,li2024enhancing,gao2024complementing,jiang2024p,yuan2024presight} show that even coarse-grained map information can provide useful supplements in structure mapping and multimodal fusion, laying a stable geometric and semantic foundation for topological reasoning.
\section{VLM and Multimodal Scene Understanding}
\label{sec:5_LLM}

With the rapid development of large language models (LLM) and vision-language models (VLM) in the field of natural language processing and multimodal understanding, autonomous driving research is ushering in a new paradigm shift from a mapping system based on geometry and topology to a more cognitive language-driven, multimodal enhanced scene understanding framework. This trend not only improves the expressiveness and interactivity of the model, but also provides a solution to the long-standing "black box problem".

\subsection{Enhanced Causal Reasoning and Interpret-Ability}

The core motivation for introducing LLMs is to improve the interpretability, interactivity, and semantic generalization capabilities of autonomous driving systems. Compared with traditional perception-control mapping methods, language models can not only predict control behaviors, but also explain their decision logic through natural language, thereby enhancing user trust and system controllability.

DriveGPT4~\cite{xu2024drivegpt4,xu2025drivegpt4} is one of the representative works in this direction. It introduces multimodal language models into end-to-end systems, supports generating driving instructions from image inputs, and provides language explanations. To support this capability, DriveGPT-4 builds a visual instruction tuning dataset for learning the mapping between driving semantics and language expressions.

DriveLM~\cite{sima2024drivelm} further simulates the multi-stage causal reasoning process of humans, proposes a graphical visual question-answering task, and decomposes the decision-making process into a structured question-answering chain, from basic perception to motion prediction, and then to behavior planning, making the model closer to human reasoning logic.

\begin{figure*}[t]
  \centering
  \includegraphics[width=\textwidth]{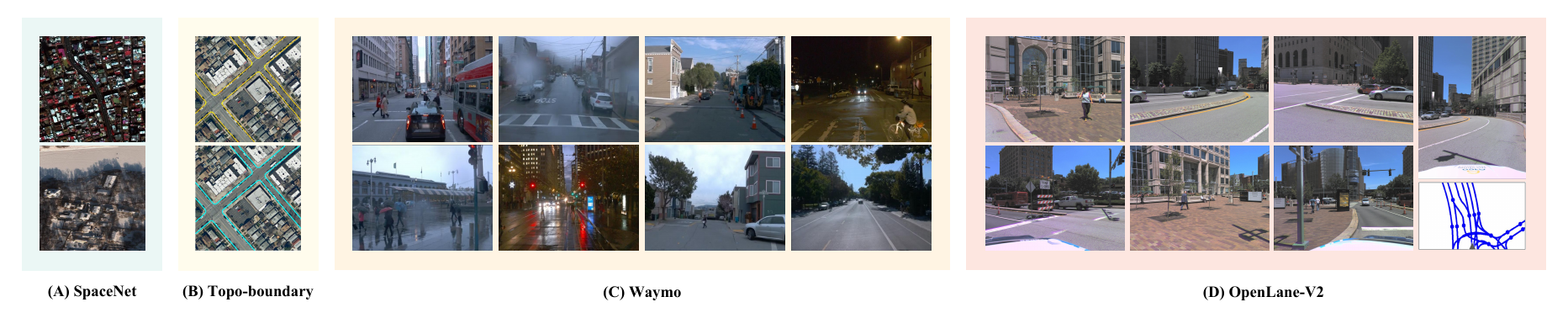} 
  \caption{Sample images from different map-related datasets for autonomous driving. Each subfigure (A)–(D) presents a distinct dataset and its corresponding visual domain. (A) SpaceNet provides high-resolution satellite imagery depicting urban layouts for tasks like overhead road and building extraction. (B) Topo-boundary includes aerial images with annotated road topology, emphasizing structured spatial relationships. (C) Waymo features diverse street-level images captured under various weather and lighting conditions, enabling robust object detection and tracking. (D) OpenLane-V2 showcases seven camera-view images captured from an autonomous vehicle, along with a bird’s-eye-view (BEV) lane graph representation, supporting fine-grained 3D lane topology understanding and multi-view perception.}
  \label{fig:datasets}
\end{figure*}

Talk2Car~\cite{deruyttere2019talk2car} is an early benchmark for exploring driving language cues. It is built on the nuScenes dataset and is used for target reference sentence understanding in scenes. However, its annotations only cover keyframes and are mostly single-target expressions. Refer-KITTI~\cite{wu2023referring} expands on the KITTI dataset to support multi-target references, which is closer to the complexity of language expression in real driving scenes. NuScenes-QA~\cite{qian2024nuscenes} opens up research on visual question answering based on driving scenes, enabling the system to answer user questions at the scene level.

Sce2DriveX~\cite{zhao2025sce2drivex}, Talk2BEV~\cite{choudhary2024talk2bev}, and LingoQA~\cite{marcu2024lingoqa} explore the deep integration of language interaction and BEV understanding, allowing users to ask tasks or queries to the system through natural language, and the system returns structured responses based on BEV scenarios. TalkBEV~\cite{choudhary2024talk2bev} and Rank2Tell~\cite{sachdeva2024rank2tell} use language to improve BEV expression and driving interpretation, further expanding the auxiliary capabilities of language.

In addition, generative world modeling research such as GAIA-2~\cite{russell2025gaia} and multimodal embedding methods ImageBind~\cite{girdhar2023imagebind} are building a unified cross-modal representation space, providing a new paradigm for the integration of perception, reasoning, and interaction for autonomous driving systems. These explorations constitute the cutting-edge research directions of multimodal understanding and language-driven causal reasoning.

\subsection{Scene-Level Interaction Modeling and Social Behavior Understanding}
Realizing human-level driving intelligence requires not only that the system understand the static environment, but also that the dynamic interactions between multiple agents need to be modeled. In recent years, research has gradually expanded from single trajectory prediction to topological understanding of social behavior.

Reasoning Multi-agent Behavioral Topology~\cite{liu2024reasoning} introduces interaction graphs to characterize complex behaviors such as yielding and overtaking, and combines structural priors with behavior labels to infer causal chains. Learning Lane Graph Representations~\cite{liang2020learning} encodes behavioral relationships in lane graphs to achieve linkage modeling of lanes and agent behaviors.

In terms of behavior prediction representation, the research has evolved from early occupancy prediction methods~\cite{hu2021fiery,hu2022st,casas2021mp3} to BEV spatial existence prediction that supports arbitrary agents~\cite{liu2023multi,kamenev2022predictionnet,li2023delving}, and achieves semantic alignment with the perception module~\cite{hu2023planning,yang2024visual}. However, fixed BEV resolution still faces occlusion problems in complex scenes, limiting the integrity of the representation. Sparse trajectory representation~\cite{lin2024eda,huang2024learning,huang2024dtpp} and joint behavior modeling~\cite{gilles2021thomas,gilles2022gohome,gu2021densetnt} alleviate this problem to a certain extent, but also introduce the challenges of modal collapse and increased computational complexity.

BeTop~\cite{liu2024reasoning} proposes to combine topological behavior probability and sparse motion pattern, and use the topological braid theory to uniformly model future interactions to improve structural stability and interpretability. Compared with implicit modeling methods such as attention or GNN~\cite{nayakanti2023wayformer,jia2023hdgt}, explicit reasoning methods such as relationship graphs~\cite{zhou2022hivt,jia2024amp} and conditional modeling are more consistent in complex interactions~\cite{rowe2023fjmp,sun2022m2i}. 

\begin{table}[t]
\scriptsize
\setlength{\tabcolsep}{4pt}
\caption{Comparison of real-world autonomous driving datasets commonly used for lane topology reasoning}
\centering
\begin{tabular}{c|c|c}
\toprule
\textbf{Dataset} & \textbf{Data Source} & \textbf{Sampling} \\
\midrule
\href{https://registry.opendata.aws/spacenet/}{SpaceNet (2018)} \cite{van2018spacenet} & Worldwide & Image \\

\href{https://github.com/JinkyuKimUCB/BDD-X-dataset}{BDD-X (2018)} \cite{kim2018textual} & Multiple US Cities & \SI{30}{Hz} \\

\href{https://www.nuscenes.org/}{nuScenes (2019)} \cite{caesar2020nuscenes} & Boston, Singapore & \SI{2}{Hz} \\

\href{https://waymo.com/open/}{Waymo (2020)} \cite{sun2020scalability} & Multiple US Cities & \SI{10}{Hz} \\

\href{https://www.argoverse.org/}{Argoverse 1  \& 2 (2019, 2023)} \cite{chang2019argoverse, wilson2023argoverse} & Miami, Pittsburgh & \SI{10}{Hz} \\

\href{https://github.com/OpenDriveLab/OpenLane-V2}{OpenLane-V2 (2023)} \cite{wang2023openlane} & Boston, Pittsburgh, Singapore & \SI{2}{Hz} \\

\href{https://github.com/TonyXuQAQ/Topo-boundary?tab=readme-ov-file}{Topo-boundary (2021)} \cite{xu2021topo} & New York City & Image \\
\bottomrule
\end{tabular}
\label{tab:dataset_summary}
\end{table}

\section{Datasets}
\label{sec:6_benchmark_dataset}

In autonomous driving research, public high-quality datasets are the key foundation for promoting algorithm development and standardized evaluation. Shown in Fig.~\ref{fig:intro} and Table.~\ref{tab:dataset_summary}, these datasets cover multiple levels from low-altitude remote sensing to on-board perception, from geometric mapping to semantic reasoning, providing solid support for map semantic understanding and structured perception research.


OpenLane-V2~\cite{wang2023openlane} is the first large-scale benchmark dataset specifically designed for lane topology reasoning. It not only supports fine-grained 3D lane detection but also includes annotations for traffic element recognition and topological structure modeling. One of its key contributions is the introduction of the OLS (OpenLane Score) index, which enables a unified evaluation framework that jointly considers geometric accuracy and topological correctness, making it a comprehensive benchmark for high-level lane understanding tasks.

NuScenes~\cite{caesar2020nuscenes} provides synchronized multi-sensor data collected from diverse urban environments under various weather and lighting conditions. It includes 360-degree LiDAR, multiple camera views, radar, GPS, and IMU data, making it a rich resource for research in object detection, tracking, sensor fusion, and multi-modal perception. Its detailed annotations and time-synchronized streams make it one of the most widely used datasets for autonomous driving perception research.

Argoverse 2~\cite{wilson2023argoverse} builds upon the original Argoverse 1~\cite{chang2019argoverse} dataset by significantly enhancing the richness of its motion trajectory data and map annotations. With high-definition semantic maps, labeled actor trajectories, and diverse urban scenes, Argoverse 2 has become a valuable resource for trajectory prediction, motion forecasting, and topological behavior modeling.

The Waymo Open Dataset~\cite{sun2020scalability} offers a vast amount of high-density LiDAR and camera data collected in complex urban scenarios across multiple cities. It is well-known for its data diversity, scale, and quality, providing strong support for the development of high-precision perception, 3D object detection, and tracking algorithms. Its recent extensions also include motion forecasting and interaction prediction benchmarks.

BDD-X~\cite{kim2018textual} is a dataset focused on image-language pairing, supporting the training and evaluation of language-guided autonomous driving systems, such as DriveGPT-4. It enables vision-and-language models to better understand driving scenarios by linking visual information with textual descriptions, which is critical for instruction following and decision reasoning.

In addition, datasets such as SpaceNet~\cite{van2018spacenet} and Topo-boundary~\cite{xu2021topo} provide high-quality annotations of road structures from aerial and satellite imagery. These datasets are particularly useful for large-scale topological mapping tasks and help to overcome the limitations of vehicle-mounted perspective datasets by offering a broader and more complete view of both urban and rural environments.

\section{Challenges and Future Outlook}
\label{sec:7_challenge_future}
\subsection{Main Challenges}

\textbf{Bottlenecks of topological consistency and structural modeling.}
Existing methods can generate geometrically accurate lane lines and map elements. However, topological consistency is still difficult to guarantee in complex scenarios. Especially in multimodal perception or end-to-end generation, topological relationships are prone to errors or omissions. This will affect path planning and decision-making. 
At the same time, it is also necessary to design loss functions and optimization mechanisms that support topology preservation for a stronger model.

\textbf{Real-time and scalability of map updates.} 
Current online map methods can quickly restore local map information. However, they lack long-term consistency. In practical applications, maps need to be continuously updated to adapt to urban changes and traffic adjustments. This places higher demands on data structure, model efficiency, and update mechanism.

\textbf{Cross-modal expression and multi-source fusion mechanism.}
With the development of large language models and multimodal technologies, autonomous driving systems have begun to have certain semantic reasoning capabilities. However, how to unify the expression of geometry, semantics, and language remains a difficult problem. Current fusion methods mostly rely on post-processing or simple splicing, and lack a unified encoding framework. In the future, more general intermediate representations, cross-modal attention mechanisms, and explainable reasoning methods are needed.

\textbf{Safety verification and high-risk behavior prediction.}
The introduction of large models increases the uncertainty of the system. Existing verification methods are difficult to cope with emerging risks. Formal verification mechanisms are not yet perfect. Extreme situations and rare behaviors are not fully modeled. The system needs to remain stable in a complex environment and be able to provide causal explanations for its output.

\textbf{Limitations of benchmarks and evaluation gaps.}
Datasets such as OpenLane-V2~\cite{wang2023openlane} and nuScenes~\cite{caesar2020nuscenes} have promoted standardization. However, they cannot cover all real-world scenarios. For example, some rare events and rule changes are still not included. Current evaluation metrics may not fully reflect the balance between map quality and safety. A new evaluation system needs to be established, including closed-loop simulation, edge scenario mining, and adversarial testing.

\subsection{Future Outlook}
\textbf{Generative World Models.}
Scene understanding in autonomous driving is evolving from discriminative models to generative world models. Unlike traditional methods that only recognize current perception content, generative models attempt to understand the causal structure and physical laws behind the environment and predict possible future states. For example, GAIA-2~\cite{russell2025gaia} achieves stronger prediction and planning capabilities by modeling environmental evolution and multi-agent interaction. This direction is expected to break the modular barriers between perception and decision-making, enabling the system to have the ability of active modeling and interactive reasoning.

\textbf{Unified reasoning engine centered on LLM.}
Future autonomous driving systems will no longer use LLMs as auxiliary modules for perception, but will upgrade them to core reasoning engines. Research represented by DriveLM~\cite{sima2024drivelm} has initially achieved that language models receive structured perception inputs and output natural language interpretations, behavioral intentions, and high-level strategies. This perception-prediction-question-answer multi-round interaction method marks a shift in the role of language models in system architecture. In the future, MLLM is expected to take on key functions such as task coordination, command translation, and semantic planning, and promote a complete semantic closed loop from BEV representation to language understanding.

\textbf{General cognitive model for embodied agents.}
Autonomous driving is a typical example of embodied intelligence, which requires the system to perform real-time perception, structural understanding, causal reasoning, and goal-driven behavior in a high-risk, dynamic physical environment. The fusion path from pixels to language not only helps to improve the comprehensive capabilities of a single system, but also provides an important exploration direction for the realization of general artificial intelligence (AGI). In future, the cognitive ability boundary of autonomous driving systems will become a key criterion for measuring the versatility and reliability of embodied intelligence systems.

\section{Conclusion}

In this survey, we provide a comprehensive review of topology perception models for autonomous driving, representing a paradigm shift from static high-definition maps to dynamic, sensor-driven representations that support real-time road semantics and topology reasoning. By analyzing key methods, we reveal the convergence of these diverse technology approaches. Despite significant progress, challenges remain, including consistent topology modeling, real-time updates, and unified cross-modal representations. 

\section{Acknowledgment}
This work was partially supported by the Key Research and Development Program of Shandong Province under Grant No. 2025CXGC010901, and by Beijing Sems Automotive Electronic Systems Co., LTD.

\clearpage
{
    \small
    \bibliographystyle{ieeenat_fullname}
    \bibliography{main}
}

\end{document}